%% file: main.tex
\documentclass[11pt,a4paper]{article}
\usepackage[hyperref]{emnlp2020}
\usepackage{times}
\usepackage{latexsym}

\usepackage{microtype}

\usepackage{amsmath}
\usepackage{amssymb}
\usepackage{graphicx}
\usepackage{rotating}

\usepackage{booktabs}

\aclfinalcopy 
\def\aclpaperid{1898} 

\newcommand{\p}[1]{P_{\textmd{MLM}}\left(#1\right)}

\newcommand{\pl}[1]{\mathcal{L}\left(#1\right)}
\newcommand{\plonly}{\mathcal{L}}
\newcommand{\pad}{\textmd{\texttt{[PAD]}}}
\newcommand{\mask}{\textmd{\texttt{[MASK]}}}
\newcommand{\stoken}{\texttt{[SOURCE]}}
\newcommand{\ttoken}{\texttt{[TARGET]}}
\newcommand{\np}{\ensuremath{n_p}}

\newcommand{\score}[1]{\textmd{Score}(#1)}

\newcommand{\sscore}[1]{\textmd{SourceScore}(#1)}
\newcommand{\sscoreonly}{SourceScore}
\newcommand{\tscore}[1]{\textmd{TargetScore}(#1)}

\newcommand{\LT}{\textsc{LaserTagger~}}
\newcommand{\FI}{\textsc{FelixInsert~}}
\newcommand{\berttobert}{\textsc{Bert2Bert~}}
\newcommand{\masker}{\textsc{Masker}}

\title{Unsupervised Text Style Transfer with Padded Masked Language Models}

\author{Eric Malmi \\
  Google Research \\
  \texttt{emalmi@google.com} \\\And
  Aliaksei Severyn \\
  Google Research \\
  \texttt{severyn@google.com} \\\And
  Sascha Rothe \\
  Google Research \\
  \texttt{rothe@google.com} \\}

\date{}

\begin{document}
\maketitle
\begin{abstract}
We propose \masker{}, an unsupervised text-editing method for style transfer. To tackle cases when no parallel source--target pairs are available, we train masked language models (MLMs) for both the source and the target domain. Then we find the text spans where the two models disagree the most in terms of likelihood. This allows us to identify the source tokens to delete to transform the source text to match the style of the target domain. 
The deleted tokens are replaced with the target MLM, and by using a \textit{padded} MLM variant, we avoid having to predetermine the number of inserted tokens.
Our experiments on sentence fusion and sentiment transfer demonstrate that \masker{} performs competitively in a fully unsupervised setting. Moreover, in low-resource settings, it improves supervised methods' accuracy by over 10 percentage points when pre-training them on silver training data generated by \masker{}.
\end{abstract}

\section{Introduction}
\label{sec:intro}

Text-editing methods \cite{EditNTS,malmi2019encode,awasthi2019parallel,mallinson2020felix}, that target monolingual sequence transduction tasks like sentence fusion, grammar correction, and text simplification, are typically more data-efficient than the traditional sequence-to-sequence methods, but they still require substantial amounts of parallel training examples to work well.
When parallel source--target training pairs are difficult to obtain, it is often still possible to collect non-parallel examples for the source and the target domain separately. For instance, negative and positive reviews can easily be collected based on the numerical review scores associated with them, which has led to a large body of work on unsupervised text style transfer, e.g., \cite{yang2018unsupervised,shen2017style,wu2019mask,li2018delete}.

The existing unsupervised style transfer methods aim at transforming a source text so that its style matches the target domain but its content stays otherwise unaltered. This is commonly achieved via text-editing performed in two steps: using one model to identify the tokens to delete and another model to infill the deleted text slots  \cite{li2018delete,xu2018unpaired,wu2019mask}. In contrast, we propose a more unified approach, showing that both of these steps can be completed using a single model, namely a \textit{masked language model} (MLM) \cite{devlin2019bert}. 
MLM is a natural choice for infilling the deleted text spans, but we can also use it to identify the tokens to delete by finding the spans where MLMs trained on the source and the target domain disagree in terms of likelihood. This is inspired by the recent observation that MLMs are effective at estimating (pseudo) likelihoods of texts \cite{wang2019bert,salazar2019masked}.
Moreover, by using a \textit{padded} variant of MLM \cite{mallinson2020felix}, we avoid having to separately model the length of the infilled text span.

To evaluate the proposed approach, \masker{}, we apply it to two tasks: sentence fusion, which requires syntactic modifications, and sentiment transfer, which requires semantic modifications. In the former case, \masker{} improves the accuracy of state-of-the-art text-editing models by more than 10 percentage points in low-resource settings by providing silver data for pretraining, while in the latter, it yields a competitive performance compared to existing unsupervised style-transfer methods.

\section{Method}
\label{sec:method}

Our approach to unsupervised style transfer is to modify source texts to match the style of the target domain. To achieve this, we can typically keep most of the source tokens and only modify a fraction of them. To determine which tokens to edit and how to edit them, we propose the following three-step approach: \\
\textit{(1)} Train padded MLMs on source domain data ($\Theta_\textmd{source}$) and on target domain data ($\Theta_\textmd{target}$). (\S\ref{sec:what}) \\
\textit{(2)} Find the text spans where the models \textit{disagree} the most to determine the tokens to delete. (\S\ref{sec:where}) \\
\textit{(3)} Use $\Theta_\textmd{target}$ to replace the deleted spans with text that fits the target domain.

\subsection{Padded Masked Language Models}
\label{sec:what}

The original MLM objective in BERT~\cite{devlin2019bert} does not model the length of infilled token spans since each \mask{} token corresponds to one wordpiece token that needs to be predicted at a given position. 
To model the length, it is possible to use an autoregressive decoder or a separate model~\cite{mansimov2019generalized}.
Instead, we use an efficient non-autoregressive \textit{padded MLM} approach by~\citet{mallinson2020felix} which enables BERT to predict \pad{} symbols when infilling a fixed-length spans of \np{} \mask{} tokens.

When creating training data for this model, spans of zero to \np{} tokens, corresponding to whole word(s), are masked out after which the mask sequences are padded to always have \np{} \mask{} tokens. For example, if $\np = 4$ and we have randomly decided to mask out tokens from $i$ to $j=i+2$ (inclusive) 
from text $W$, the corresponding input sequence is:
\begin{equation*}
\begin{split}
W_{\backslash{i:j}} = ( & w_1, \ldots, w_{i-1}, \mask, \mask, \\
& \mask, \mask, w_{i+3}, \ldots, w_{|W|}).
\end{split}
\end{equation*}
The targets for the first three \mask{} tokens are the original masked out tokens, i.e. $w_{i}, w_{i+1}, w_{i+2}$, while for the remaining token the model is trained to output a special \pad{} token.

Similar to \cite{wang2019bert,salazar2019masked}, we can compute the pseudo-likelihood ($\plonly$) of the original tokens $W_{i:j}$ according to:
\begin{align*}
   \pl{W_{i:j} \mid W_{\backslash{i:j}}; \Theta} = \prod_{t=i}^{j} \p{w_t \mid W_{\backslash{i:j}}; \Theta} \\
   \times \prod_{t=j+1}^{i + \np - 1} \p{\pad_t \mid W_{\backslash{i:j}}; \Theta},
\end{align*}
where $\p{*_t \mid W_{\backslash{i:j}}; \Theta}$ denotes the probability of the random variable corresponding to the $t$-th token in $W_{\backslash{i:j}}$ taking value $w_t$ or \pad. 
Furhermore, we can compute the maximum pseudo-likelihood infilled tokens $\widehat{W}_{i:j} = \arg \max_{W_{i:j}} \pl{W_{i:j} \mid W_{\backslash{i:j}}; \Theta}$ by taking the most likely insertion for each \mask{} independently, as done by the regular BERT. These maximum likelihood estimates are used both when deciding which spans to edit (as described in \S\ref{sec:where}) as well as when replacing the edited spans.

In practice, instead of training two separate models for the source and target domain, we train a single conditional model. Conditioning on a domain is achieved by prepending a special token (\stoken{} or \ttoken{}) to each token sequence fed to the model.\footnote{The motivation for using a joint model instead of two separate models is to share model weights to give more consistent likelihood estimates. An alternative way of conditioning the model would be to add a domain embedding to each token embedding as proposed by \citet{wu2019mask}.}
At inference time, padded MLM can decide to insert zero tokens (by predicting \pad{} for each mask) or up to \np{} tokens based on the bidirectional context it observes.
In our experiments, we set $\np=4$.\footnote{In early experiments, we also tested $\np=8$, but this resulted in fewer grammatical predictions since each token is predicted independently. To improve the predictions, we could use SpanBERT~\cite{joshi2020spanbert}, which is designed to infill spans, or an autoregressive model like T5~\cite{raffel2019exploring}.}

\subsection{Where to edit?}
\label{sec:where}

Our approach to using MLMs to determine where to delete and insert tokens is to find text spans where the source and target model disagree the most. Here we introduce a scoring function to quantify the level of disagreement.

First, we note that any span of source tokens that has a low likelihood in the target domain is a candidate span to be replaced or deleted. That is, source tokens from index $i$ to $j$ should be more likely to be deleted the lower the likelihood $\pl{W_{i:j} \mid W_{\backslash{i:j}}; \Theta_\textmd{target}}$ is.
Moreover, if two spans have equally low likelihoods under the target model, but one of them has a higher maximum likelihood replacement $\widehat{W}_{i:j}^\textmd{target}$, then it is safer to replace the latter. For example, if a sentiment transfer model encounters a polarized word of the wrong sentiment and an arbitrary phone number, it might evaluate both of them as unlikely. However, the model will be more confident about how to replace the polarized word, so it should try to replace that rather than the phone number. Thus the first component of our scoring function is:
\[
\begin{split}
\tscore{i,j} = \; & \pl{\widehat{W}_{i:j}^\textmd{target} \mid W_{\backslash{i:j}}; \Theta_\textmd{target}} \\
   & - \pl{W_{i:j} \mid W_{\backslash{i:j}}; \Theta_\textmd{target}}.
\end{split}
\]
This function can be used on its own without having access to a source domain corpus, but in some cases, this leads to undesired replacements. The target model can be very confident that, e.g., a rarely mentioned entity should be replaced with a more common entity, 
although this type of edit does not help with transferring the style of the source text toward the target domain. To address this issue, we introduce a second scoring component leveraging the source domain MLM:
\[
{\small
\begin{aligned}
  \sscore{i,j} = - \max \Big[ 0, \pl{\widehat{W}_{i:j}^\textmd{target} \mid W_{\backslash{i:j}}; \Theta_\textmd{source}} \\
  - \pl{W_{i:j} \mid W_{\backslash{i:j}}; \Theta_\textmd{source}} \Big]
\end{aligned}
}
\]
By adding this component to $\tscore{i,j}$, we can counter for edits that only increase the likelihood of a span under $\Theta_\textmd{target}$ but do not push the style closer to the target domain.\footnote{\sscore{i,j} is capped at zero to prevent it from dominating the overall score. Otherwise, we might obtain low-quality edits in cases where the likelihood of the source span $W_{i:j}$ is high under the source model and low under the target model but no good replacements exist according to the target model. Given the lack of good replacements, $\widehat{W}_{i:j}^\textmd{target}$ may end up being ungrammatical, pushing \sscore{i,j} close to $1$ and thus making it a likely edit, although \tscore{i,j} remains low.}

Our overall scoring function is given by:
\[
{\small
\score{i,j} = \tscore{i,j} + \sscore{i,j}.
}
\]
To determine the span to edit, we compute $\arg \max_{i,j} \score{i,j}$,
where $1 \leq i \leq |W| + 1$ and $i - 1 \leq j \leq i + \np - 1$. The case $j = i-1$ denotes an empty source span, meaning that the model does not delete any source tokens but only adds text before the $i$-th source token.

\begin{figure}[tb]
\centering
\includegraphics[width=\columnwidth]{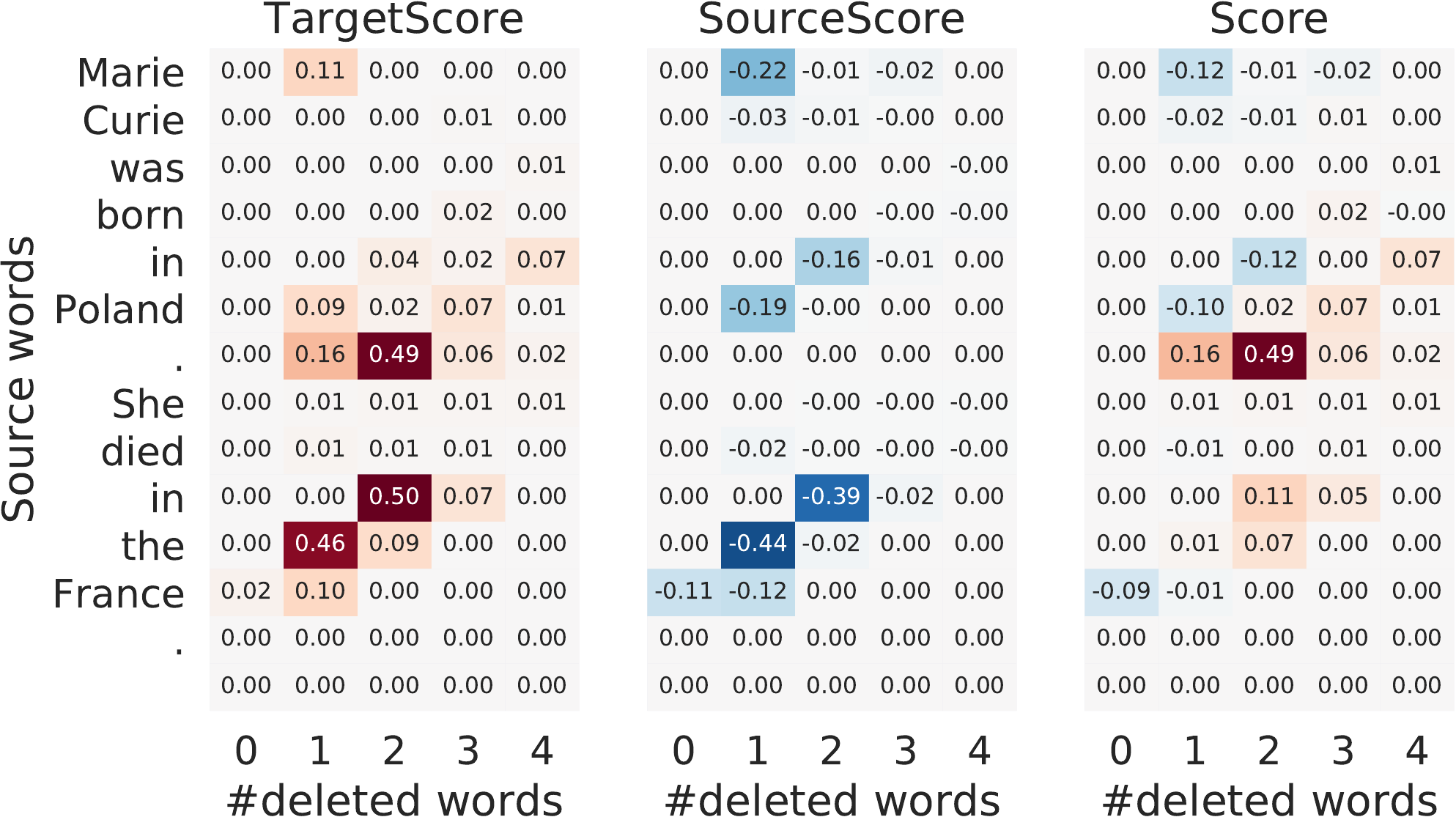}
\caption{\masker{} replaces span ``\textit{. She}'' by ``\textit{and \pad{} \pad{} \pad{}}'', resulting in the following fused sentence: \textit{Marie Curie was born in Poland and died in the France .} 
}
\label{fig:scores}
\end{figure}

The process for selecting the span to edit is illustrated in Figure~\ref{fig:scores}, where the source text corresponds to two sentences to be fused. The source MLM has been trained on unfused sentences and the target MLM on fused sentences from the DiscoFuse corpus~\cite{geva2019discofuse}. In this example, the target model is confident that either the boundary between the two sentences or the grammatical mistake ``\textit{in {\color{red}the} France}'' should be edited. However, also the source model is confident that the grammatical mistake should be edited, so the model correctly ends up editing the words ``\textit{. She}'' at the sentence boundary. The resulting fused sentence is: \textit{Marie Curie was born in Poland and died in the France .}

\paragraph{Efficiency.}
The above method is computationally expensive since producing a single edit requires $\mathcal{O}(|W| \times \np)$ BERT inference steps -- although these can be run in parallel. The model can be distilled into a much more efficient supervised student model without losing -- and even gaining -- accuracy as shown in our experiments.
This is done by applying \masker{} to the unaligned source and target examples to generate aligned silver data for training the student model.

\section{Experiments}
\label{sec:experiments}

We evaluate \masker{} on two different types of tasks: \textit{sentence fusion} and \textit{sentiment transfer}. 
For both experiments, we only apply \masker{} once to edit a single span of at most four tokens, since the required edits are often local.\footnote{We tried running multiple iterations of \masker, but this somewhat decreased the accuracy of the method. When parallel development data is available, it could potentially be used to optimize a threshold of $\score{i, j}$ so that the model could be called repeatedly until the $\score{i, j}$ falls below the threshold. Alternatively, it would be interesting to explore methods for simultaneously identifying multiple, not necessarily adjacent, spans to edit.}

\subsection{Sentence Fusion}
\label{sec:fusion}

Sentence fusion is the task of fusing two (or more) incoherent input sentences into a single coherent sentence or paragraph, and DiscoFuse~\cite{geva2019discofuse} is a recent parallel dataset for sentence fusion. We study both a fully unsupervised setting as well as a low-resource setting.

\paragraph{Unsupervised.}
First, we remove the alignment between unfused and fused examples in the training set of 4.5 million examples and finetune \masker{} on the resulting, non-parallel source and target corpora. This model yields an Exact match accuracy (which is a standard metric for sentence fusion \cite{geva2019discofuse,rothe2019leveraging}) of 12.65 on the development set. This is already on par with the Exact score of 12.32 obtained by a supervised \LT model \cite{malmi2019encode} trained on 450 examples.
We also test ablating \sscoreonly{}, which results in a decreased Exact score of 2.14, attesting to the importance of using the source model.
Finally, we test our model on the reverse direction of going from a fused text to unfused sentences. Here, \masker{} yields a significantly higher Exact score of 23.18. This direction is typically easier since it does not involve predicting discourse markers, which would require modeling the semantic relation between two sentences. The predictions of the reverse model are used in the low-resource experiments. 
The unsupervised results are summarized in Table~\ref{tab:unsupervised_discofuse}.

\begin{table}[tb]
\centering
{\small
\begin{tabular}{lc}
\toprule
\textbf{Method} & \textbf{Exact score} \\
\midrule
\masker{} (unfused $\rightarrow$ fused) & 12.65 \\
\quad \textit{ablating} SourceScore & 2.14 \\
\midrule
\masker{} (fused $\rightarrow$ unfused) & 23.18 \\
\bottomrule
\end{tabular}
}
\caption{\label{tab:unsupervised_discofuse} Unsupervised sentence fusion results.}
\end{table}

\paragraph{Low resource.}
We use \masker{} to generate noisy unfused sentences for 46K target fusions in the DiscoFuse development set. This silver data is used to pretrain three different model architectures, \LT~\cite{malmi2019encode}, \FI~\cite{mallinson2020felix}, and \berttobert~\cite{rothe2019leveraging}, which have previously been used for training fusion models under low-resource settings. 
The results on the test set (45K examples) without and with pretraining on \masker{} outputs are shown in Table~\ref{tab:discofuse}. On average, the silver data from \masker{} improves the Exact score by 13.37 when 450 parallel training examples are available and still by 2.01 when 45\,000 parallel examples are available.

\begin{table}[tb]
\centering
\resizebox{\columnwidth}{!}{%
\begin{tabular}{lcccc}
\toprule
\textbf{Method} & \multicolumn{4}{c}{\textbf{Parallel training examples}} \\
 & \textbf{0} & \textbf{450} & \textbf{4500} & \textbf{45000} \\
\midrule
\LT & 0.00 & 12.32 & 25.74 & 38.46 \\
+ \masker{} silver data & 19.61 &	25.97 &	34.20 &	42.41 \\
\midrule
\FI & 0.00 &	15.34 &	34.11 &	46.09 \\
+ \masker{} silver data & 18.22	& 25.23  & 38.43 & 47.21 \\
\midrule
\berttobert &  0.00 &	0.00 &	3.35 &	42.07  \\
+ \masker{} silver data &  13.05 & 16.57 & 30.14 & 43.03 \\
\toprule
\textbf{Average improvement} & \textbf{16.96} &	\textbf{13.37} &	\textbf{13.19} &	\textbf{2.01} \\
\bottomrule
\end{tabular}
}
\caption{\label{tab:discofuse} Low-resource sentence fusion results.
Using the predictions of \masker{} as silver data to pretrain models improves the Exact score.
}
\end{table}

\subsection{Sentiment Transfer}
\label{sec:sentiment}

In sentiment transfer, the task is to change a text's sentiment from negative to positive or vice versa. We use a dataset of Yelp reviews \cite{li2018delete}, containing 450K training, 4K development, and 1K test examples. Half of the test reviews are positive and half negative, and human annotators have written a reference review of the opposite sentiment for each test review. We use the same automatic evaluation metrics used in previous work: \textit{BLEU} score 
and \textit{accuracy} that a classifier trained to distinguish negative and positive reviews assigns to the modified reviews being of the target sentiment.

We finetune the MLMs on the training set and apply the resulting \masker{} model to the test set. Additionally, we apply the \masker{} model to the non-parallel training set to create parallel silver data and train a \LT model. Interestingly, the latter setup outperforms \masker{} alone (15.3 vs. 14.5 BLEU score; 49.6 vs. 40.9 sentiment accuracy). We think this happens because \LT employs a restricted vocabulary of 500 most frequently inserted phrases, which prevents the model from reproducing every spurious infilling that the padded MLM may have produced, effectively regularizing \masker{}. In Table~\ref{tab:yelp}, we report these results along with baseline methods developed specifically for the sentiment transfer task by \citet{li2018delete} and \citet{wu2019mask}. Overall, \masker{} yields a competitive performance although \textsc{AC-MLM} w/ attention-based \cite{wu2019mask} slightly outperforms it.

\begin{table}[tb]
\centering
\resizebox{\columnwidth}{!}{%
\begin{tabular}{lcc}
\toprule
\textbf{Method} & \textbf{BLEU} & \textbf{ACC (\%)} \\
\midrule
\textsc{DeleteAndRetrieval} &	8.5 & \textbf{87.9} \\
\midrule
\textsc{AC-MLM} w/ frequency-ratio &	13.2 & 37.9 \\
\textsc{AC-MLM} w/ attention-based &	\textbf{15.7} & 53.4 \\
\textsc{AC-MLM} w/ fusion-method &	15.3 & 40.9 \\
\midrule
\masker{} &	14.5 & 40.9 \\
\LT w/ \masker{} silver data &	15.3 & 49.6 \\
\bottomrule
\hline
\end{tabular}
}
\caption{\label{tab:yelp} Yelp review sentiment transfer results.}
\end{table}

\section{Related Work}

Section~\ref{sec:intro} provides a high-level overview of the related work. Closest to this work is the \textsc{AC-MLM} sentiment transfer method by \citet{wu2019mask}. 
This method first identifies the tokens to edit based on $n$-gram frequencies in the source vs. target domain (as proposed by \citet{li2018delete}) and based on LSTM attention scores (as proposed by \citet{xu2018unpaired}). Then it replaces the edited tokens using a conditional MLM. In contrast to their work, our approach leverages the same MLM for both identifying the (possibly empty) span of tokens to edit and for infilling the deleted span. Moreover, our padded MLM determines the number of tokens to insert without having to pre-specify it. In that sense, it is similar to the recently proposed Blank Language Model \cite{shen2020blank}.

In addition to the two applications studied in this work, it would be interesting to evaluate \masker{} on other style transfer tasks. Tasks for which unsupervised methods have recently been developed include formality transfer \cite{rao2018dear,luo2019dual}, lyrics style transfer \cite{nikolov2020conditional,lee2019neural}, text simplification \cite{paetzold2016unsupervised,surya2019unsupervised}, and sarcasm generation \cite{mishra2019modular,chakrabarty2020r}.

\section{Conclusions}

We have introduced a novel way of using masked language models for text-editing tasks where no parallel data is available. The method is based on training an MLM for source and target domains, identifying the tokens to delete by finding the spans where the two models disagree in terms of likelihood, and infilling more appropriate text with the target MLM. This approach yields a competitive performance in fully unsupervised settings and substantially improves over previous works in low-resource settings.

\bibliographystyle{acl_natbib}
\bibliography{emnlp2020}

\appendix

\newpage

\section{Examples of Model Outputs}

To further illustrate how \masker{} works, Table~\ref{tab:fig1} shows all the input sequences and the output scores that go into computing Figure 1 in the main paper. Furthermore, Tables~\ref{tab:df_appendix}~and~\ref{tab:yelp_appendix} present random samples of correct and incorrect outputs by \masker{} for the DiscoFuse and Yelp datasets.

\section{Hyperparameter Settings}
\label{sec:experiments_appendix}

We did not perform any hyperparameter tuning, but used a fixed learning rate of 3e-5 and a batch size roughly proportionate to the training set size (see Table~\ref{tab:hyperparams} for the chosen values). The number of training steps was determined by running the training until convergence and choosing the checkpoint with the highest validation score, shown in Table~\ref{tab:hyperparams}.

\begin{table*}[tb]
\centering
\resizebox{\textwidth}{!}{%
\begin{tabular}{llccc}
\toprule
\textbf{Method} & \textbf{Dataset} & \textbf{Learning rate} & \textbf{Batch size} & \textbf{Exact score (validation)} \\
\midrule
Padded MLM & DiscoFuse & 3e-5 & 512 & 44.03 \\
\LT & DiscoFuse, \masker{} silver data & 3e-5 & 256 & 27.09 \\
\LT & DiscoFuse, 450 & 3e-5 & 32 & 33.16 \\
\LT & DiscoFuse, 4500 & 3e-5 & 64 & 43.36 \\
\LT & DiscoFuse, 45000 & 3e-5 & 128 & 49.43 \\
Padded MLM & Yelp & 3e-5 & 2048 & 49.15 \\
\LT & Yelp, \masker{} silver data (neg to pos) & 3e-5 & 512 & 31.79 \\
\LT & Yelp, \masker{} silver data (pos to neg) & 3e-5 & 512 & 31.15 \\
\bottomrule
\hline
\end{tabular}
}
\caption{\label{tab:hyperparams} Hyperparameter settings for the proposed method in Table 1 and 2, along with the Exact scores on validation set. For Padded MLM, the validation score refers to the accuracy of predicting all four masked tokens correctly.}
\end{table*}

\section{Other Experimental Details}

\paragraph{Code.} The padded MLM implementation is based on:
\url{https://github.com/google-research/bert}.
\LT{} code is available at: \url{https://github.com/google-research/lasertagger}

\paragraph{Datasets.} The DiscoFuse dataset \cite{geva2019discofuse} is available at: \url{https://github.com/google-research-datasets/discofuse}. The Yelp review dataset \cite{li2018delete} is available at: \url{https://github.com/lijuncen/Sentiment-and-Style-Transfer}.

\paragraph{Evaluation.} To compute BLEU scores, we used the implementation of \citet{wu2019mask}: \url{https://github.com/IIEKES/MLM_transfer}. The AC-MLM baseline predictions after 10 training epochs are taken from the directory. For the sentiment classification accuracy score, we trained a BERT model, which yields an accuracy of 98.4\% on the development set (slightly higher than the CNN classifier used by \citet{shen2020blank} which has an accuracy of 97.7\%). The Exact scores reported in the paper were computed after lowercasing the predictions and the targets.

\paragraph{Padded MLM pretraining.} The padded masked language model used in our experiments uses the uncased BERT-base architecture \cite{devlin2019bert} with 110M parameters. It is pretrained with the maximum pad length of $\np = 4$ on the Wikipedia and books corpora that the original BERT was also trained on. 
When creating MLM finetuning data for the source and the target domain, we always mask out only a single span of zero to four input tokens so that the masked span corresponds to whole word(s). The accuracy of the MLM at filling the masked span correctly is 44\% for sentence fusion and 49\% for sentiment transfer as shown in Table~\ref{tab:hyperparams}.

\paragraph{Computing infrastructure.} The models were trained using Tensor Processing Units (TPUs). Inference was distributed to multiple CPUs using Apache Beam and Google Cloud.

\paragraph{Runtime.} Inference time increases with the sequence length. For the example in Figure 1 of the main paper, prediction takes 52 seconds when running BERT inference on CPU. Using GPUs or TPUs can significantly reduce the runtime, but we chose to use CPUs to be able to distribute the computation more effectively. Moreover, after distilling the model into a \LT{} model (the autoregressive variant) as done in our experiments, inference takes only 535 milliseconds on GPU~\cite{malmi2019encode}.

\input{discofuse_outputs.tex}

\input{yelp_outputs.tex}

\newpage

\input{fig1_table.tex}

\end{document}

%% file: discofuse_outputs.tex
\begin{table*}[tb]
\centering
\resizebox{\textwidth}{!}{%
\begin{tabular}{ll}

\toprule
\multicolumn{2}{l}{\textbf{Random Sample of \textcolor{blue}{Correct} \masker{} Predictions}} \\
\midrule
Source     & the boat was hoisted aboard the carpathia along with other titanic lifeboats . the boat was brought to new york . \\
Prediction & the boat was hoisted aboard the carpathia along with other titanic lifeboats and brought to new york . \\
\midrule
Source     & beausoleil was a good - looking and rebellious youth . by 15 , beausoleil was sent to reform school . \\
Prediction & beausoleil was a good - looking and rebellious youth . by 15 , he was sent to reform school . \\
\midrule
Source     & it is believed that in terms of antiquity , this temple pre-dates the srirangam temple , . the name aadi vellarai . \\
Prediction & it is believed that in terms of antiquity , this temple pre-dates the srirangam temple , hence the name aadi vellarai . \\
\midrule
Source     & john was in charge of the roads north of kapunda . ben had yorke peninsula and the southern routes . \\
Prediction & john was in charge of the roads north of kapunda , while ben had yorke peninsula and the southern routes . \\
\midrule
Source     & in early 2018 , the central bank re-released the l - qiaif regime . it could replicate the section 110 spv . \\
Prediction & in early 2018 , the central bank re-released the l - qiaif regime so that it could replicate the section 110 spv . \\
\midrule
Source     & he also set up trade schools . girls could earn their living . \\
Prediction & he also set up trade schools so that girls could earn their living . \\
\midrule
Source     & the prime minister supplied the reason why : she said it was . we depend for our liberty and freedom upon the united states . \\
Prediction & the prime minister supplied the reason why : she said it was because we depend for our liberty and freedom upon the united states . \\
\midrule
Source     & long - distance routes are listed here . those entirely or mostly within one vdot district are at the following pages . \\
Prediction & long - distance routes are listed here , while those entirely or mostly within one vdot district are at the following pages . \\
\bottomrule
\multicolumn{2}{c}{} \\
\toprule
\multicolumn{2}{l}{\textbf{Random Sample of \textcolor{red}{Incorrect} \masker{} Predictions}} \\
\midrule
Source     & she claimed that she did not consult the party . she believed the outcome had been assured . \\
Prediction & she claimed that she did not consult the party . however , she believed the outcome had been assured . \\
Target     & she claimed that she did not consult the party because she believed the outcome had been assured . \\
\midrule
Source     & the 1927 german football championship was won by 1 . the 1927 german football championship is the 20th edition of the competition . \\
Prediction & the 1927 german football championship was won by 1 . the 1927 german football championship was the 20th edition of the competition . \\
Target     & the 1927 german football championship , the 20th edition of the competition , was won by 1 . \\
\midrule
Source     & many other fourteeners surround elbert in all directions . it is very close to central colorado 's collegiate peaks . \\
Prediction & many other fourteeners surround elbert because it is very close to central colorado 's collegiate peaks . \\
Target     & many other fourteeners surround elbert in all directions , and it is very close to central colorado 's collegiate peaks . \\
\midrule
Source     & the others want to return to centralville . davidson orders them back to fight the athsheans . \\
Prediction & the others want to return to centralville , but davidson orders them back to fight the athsheans . \\
Target     & although the others want to return to centralville , davidson orders them back to fight the athsheans . \\
\midrule
Source     & in march 2015 , she suffered a brain aneurysm . in march 2015 , she is expected to fully recover after surgery in the summer of 2015 . \\
Prediction & in march 2015 , she suffered a brain aneurysm . however , she is expected to fully recover after surgery in the summer of 2015 . \\
Target     & in march 2015 , she suffered a brain aneurysm but is expected to fully recover after surgery in the summer of 2015 . \\
\midrule
Source     & johann adam weishaupt died in gotha in 1830 . johann adam weishaupt is the founder of the order of the illuminati . \\
Prediction & johann adam weishaupt died in gotha in 1830 . johann weishashaupt is the founder of the order of the illuminati . \\
Target     & johann adam weishaupt , the founder of the order of the illuminati , died in gotha in 1830 . \\
\midrule
Source     & ohio state senator . ohio state served a six - year term as a justice of the ohio supreme court . \\
Prediction & ohio state senator . ohio secretary of state served a six - year term as a justice of the ohio supreme court . \\
Target     & ohio state senator and served a six - year term as a justice of the ohio supreme court . \\
\midrule
Source     & as minister he was responsible for this event . there was considerable controversy surrounding the disaster he did not step down . \\
Prediction & as minister he was responsible for this event . however , there was considerable controversy surrounding the disaster he did not step down . \\
Target     & as minister he was responsible for this event , although there was considerable controversy surrounding the disaster he did not step down . \\
\bottomrule

\end{tabular}
}
\caption{\label{tab:df_appendix} Random samples of correct and incorrect prediction on the DiscoFuse development set by the \masker{} model going from unfused to fused sentences (Exact score 12.65).}
\end{table*}

%% file: yelp_outputs.tex
\begin{table*}[tb]
\centering
{\small
\begin{tabular}{ll}

\toprule
\multicolumn{2}{l}{\textbf{Random Sample of \textcolor{blue}{Correct} \masker{} Predictions}} \\
\midrule
Source     & so far i 'm not really impressed . \\
Prediction & so far i 'm really impressed . \\
\midrule
Source     & either way i would never recommend buying from camping world . \\
Prediction & either way i would recommend buying from camping world . \\
\midrule
Source     & this is a horrible venue . \\
Prediction & this is a great venue . \\
\midrule
Source     & this place is a terrible place to live ! \\
Prediction & this place is a great place to live ! \\
\midrule
Source     & i 'm not one of the corn people . \\
Prediction & i 'm one of the corn people . \\
\midrule
Source     & this is easily the worst greek food i 've had in my life . \\
Prediction & this is easily the best greek food i 've had in my life . \\
\midrule
Source     & the sandwich was not that great . \\
Prediction & the sandwich was great . \\
\midrule
Source     & its also not a very clean park . \\
Prediction & its also a very clean park . \\
\bottomrule
\multicolumn{2}{c}{} \\
\toprule
\multicolumn{2}{l}{\textbf{Random Sample of \textcolor{red}{Incorrect} \masker{} Predictions}} \\
\midrule
Source     & also , could they not bring a single pack of cheese or red peppers ? \\
Prediction & also , could they bring a single pack of cheese or red peppers ? \\
Target     & they had plenty of cheese packets and red pepper. \\
\midrule
Source     & service was average but could not make up for the poor food and drink . \\
Prediction & service was good but could not make up for the poor food and drink . \\
Target     & service was above average as well as the food and drink . \\
\midrule
Source     & the only saving grace was the black beans . \\
Prediction & the saving grace was the black beans . \\
Target     & one of several saving graces was the black beans \\
\midrule
Source     & the rest of their food is edible but their employees and service are horrible . \\
Prediction & the rest of their food is edible and their employees and service are horrible . \\
Target     & the food is great but the employees werent moving fast enough \\
\midrule
Source     & crab cakes , salt , no other flavor . \\
Prediction & crab cakes , salt , no other . \\
Target     & crab cakes were salty and delicious. \\
\midrule
Source     & i was very disappointed with this place . \\
Prediction & i love this place . \\
Target     & i wasn't disappointed with this place at all. \\
\midrule
Source     & my pad thai tasted like thai rice noodles with barbeque sauce . \\
Prediction & the thai tasted like thai rice noodles with barbeque sauce . \\
Target     & the pad thai tasted like the best i have had in a while \\
\midrule
Source     & they also have lost sight of what good deli food is . \\
Prediction & they also have lost sight of how good deli food is . \\
Target     & the know what good deli food is \\
\bottomrule

\end{tabular}
}
\caption{\label{tab:yelp_appendix} Random samples of correct and incorrect prediction on the Yelp review test set by the \masker{} model going from negative to positive reviews.}
\end{table*}

%% file: fig1_table.tex
\begin{sidewaystable*}
\centering
{\scriptsize
\begin{tabular}{ccllccccccc}
\toprule
$i$ & $j$ & Masked input $W_{\backslash i:j}$ & Replacement $\widehat{W}_{i:j}^\textmd{target}$ & TargetScore & SourceScore & Score & $\mathcal{L}_1$ & $\mathcal{L}_2$ & $\mathcal{L}_3$ & $\mathcal{L}_4$ \\
\midrule
0 & 0 & [MASK] [MASK] [MASK] [MASK] marie cu \#\#rie was born in poland . she died in the france . & [PAD] [PAD] [PAD] [PAD] & 0.000 & 0.000 & 0.000 & 0.421 & 0.421 & 0.491 & 0.491 \\
0 & 1 & [MASK] [MASK] [MASK] [MASK] cu \#\#rie was born in poland . she died in the france . & [PAD] [PAD] [PAD] [PAD] & 0.108 & -0.225 & -0.117 & 0.116 & 0.007 & 0.234 & 0.009 \\
0 & 2 & [MASK] [MASK] [MASK] [MASK] was born in poland . she died in the france . & z [PAD] [PAD] [PAD] & 0.001 & -0.006 & -0.005 & 0.001 & 0.000 & 0.006 & 0.000 \\
0 & 3 & [MASK] [MASK] [MASK] [MASK] born in poland . she died in the france . & she was was [PAD] & 0.000 & -0.021 & -0.021 & 0.000 & 0.000 & 0.021 & 0.000 \\
1 & 0 & marie [MASK] [MASK] [MASK] [MASK] cu \#\#rie was born in poland . she died in the france . & [PAD] [PAD] [PAD] [PAD] & 0.000 & 0.000 & 0.000 & 0.357 & 0.357 & 0.324 & 0.324 \\
1 & 1 & marie [MASK] [MASK] [MASK] [MASK] was born in poland . she died in the france . & [PAD] [PAD] [PAD] [PAD] & 0.003 & -0.025 & -0.023 & 0.003 & 0.000 & 0.025 & 0.000 \\
1 & 2 & marie [MASK] [MASK] [MASK] [MASK] born in poland . she died in the france . & was [PAD] was [PAD] & 0.000 & -0.011 & -0.011 & 0.000 & 0.000 & 0.011 & 0.000 \\
1 & 3 & marie [MASK] [MASK] [MASK] [MASK] in poland . she died in the france . & was born born born & 0.007 & -0.002 & 0.006 & 0.007 & 0.000 & 0.002 & 0.000 \\
2 & 0 & marie cu \#\#rie [MASK] [MASK] [MASK] [MASK] was born in poland . she died in the france . & [PAD] [PAD] [PAD] [PAD] & 0.000 & 0.000 & 0.000 & 0.624 & 0.624 & 0.708 & 0.708 \\
2 & 1 & marie cu \#\#rie [MASK] [MASK] [MASK] [MASK] born in poland . she died in the france . & was [PAD] [PAD] [PAD] & 0.000 & 0.000 & 0.000 & 0.595 & 0.595 & 0.684 & 0.684 \\
2 & 2 & marie cu \#\#rie [MASK] [MASK] [MASK] [MASK] in poland . she died in the france . & was born [PAD] [PAD] & 0.000 & 0.000 & 0.000 & 0.567 & 0.567 & 0.412 & 0.412 \\
2 & 3 & marie cu \#\#rie [MASK] [MASK] [MASK] [MASK] poland . she died in the france . & was born in [PAD] & 0.000 & 0.000 & 0.000 & 0.208 & 0.208 & 0.269 & 0.269 \\
2 & 4 & marie cu \#\#rie [MASK] [MASK] [MASK] [MASK] . she died in the france . & was a [PAD] [PAD] & 0.009 & -0.001 & 0.007 & 0.009 & 0.000 & 0.001 & 0.000 \\
3 & 0 & marie cu \#\#rie was [MASK] [MASK] [MASK] [MASK] born in poland . she died in the france . & [PAD] [PAD] [PAD] [PAD] & 0.000 & 0.000 & 0.000 & 0.983 & 0.983 & 0.866 & 0.866 \\
3 & 1 & marie cu \#\#rie was [MASK] [MASK] [MASK] [MASK] in poland . she died in the france . & born [PAD] [PAD] [PAD] & 0.000 & 0.000 & 0.000 & 0.884 & 0.884 & 0.847 & 0.847 \\
3 & 2 & marie cu \#\#rie was [MASK] [MASK] [MASK] [MASK] poland . she died in the france . & born in [PAD] [PAD] & 0.000 & 0.000 & 0.000 & 0.477 & 0.477 & 0.476 & 0.476 \\
3 & 3 & marie cu \#\#rie was [MASK] [MASK] [MASK] [MASK] . she died in the france . & born french [PAD] [PAD] & 0.018 & -0.000 & 0.018 & 0.018 & 0.001 & 0.001 & 0.000 \\
3 & 4 & marie cu \#\#rie was [MASK] [MASK] [MASK] [MASK] she died in the france . & born in [PAD] [PAD] & 0.004 & -0.004 & -0.000 & 0.004 & 0.000 & 0.004 & 0.000 \\
4 & 0 & marie cu \#\#rie was born [MASK] [MASK] [MASK] [MASK] in poland . she died in the france . & [PAD] [PAD] [PAD] [PAD] & 0.000 & 0.000 & 0.000 & 0.863 & 0.863 & 0.958 & 0.958 \\
4 & 1 & marie cu \#\#rie was born [MASK] [MASK] [MASK] [MASK] poland . she died in the france . & in [PAD] [PAD] [PAD] & 0.000 & 0.000 & 0.000 & 0.601 & 0.601 & 0.717 & 0.717 \\
4 & 2 & marie cu \#\#rie was born [MASK] [MASK] [MASK] [MASK] . she died in the france . & in the [PAD] [PAD] & 0.038 & -0.158 & -0.120 & 0.042 & 0.004 & 0.164 & 0.006 \\
4 & 3 & marie cu \#\#rie was born [MASK] [MASK] [MASK] [MASK] she died in the france . & in [PAD] [PAD] [PAD] & 0.017 & -0.012 & 0.005 & 0.017 & 0.001 & 0.016 & 0.004 \\
4 & 4 & marie cu \#\#rie was born [MASK] [MASK] [MASK] [MASK] died in the france . & in [PAD] and [PAD] & 0.069 & 0.000 & 0.069 & 0.069 & 0.000 & 0.000 & 0.000 \\
5 & 0 & marie cu \#\#rie was born in [MASK] [MASK] [MASK] [MASK] poland . she died in the france . & [PAD] [PAD] [PAD] [PAD] & 0.000 & 0.000 & 0.000 & 0.466 & 0.466 & 0.698 & 0.698 \\
5 & 1 & marie cu \#\#rie was born in [MASK] [MASK] [MASK] [MASK] . she died in the france . & the [PAD] [PAD] [PAD] & 0.094 & -0.192 & -0.098 & 0.097 & 0.003 & 0.196 & 0.004 \\
5 & 2 & marie cu \#\#rie was born in [MASK] [MASK] [MASK] [MASK] she died in the france . & the , [PAD] [PAD] & 0.022 & -0.002 & 0.021 & 0.023 & 0.001 & 0.005 & 0.003 \\
5 & 3 & marie cu \#\#rie was born in [MASK] [MASK] [MASK] [MASK] died in the france . & the and [PAD] [PAD] & 0.069 & 0.000 & 0.069 & 0.069 & 0.000 & 0.000 & 0.000 \\
5 & 4 & marie cu \#\#rie was born in [MASK] [MASK] [MASK] [MASK] in the france . & germany but grew [PAD] & 0.008 & 0.000 & 0.008 & 0.008 & 0.000 & 0.000 & 0.000 \\
6 & 0 & marie cu \#\#rie was born in poland [MASK] [MASK] [MASK] [MASK] . she died in the france . & [PAD] [PAD] [PAD] [PAD] & 0.000 & 0.000 & 0.000 & 0.763 & 0.763 & 0.894 & 0.894 \\
6 & 1 & marie cu \#\#rie was born in poland [MASK] [MASK] [MASK] [MASK] she died in the france . & , [PAD] [PAD] [PAD] & 0.161 & 0.000 & 0.161 & 0.256 & 0.095 & 0.009 & 0.823 \\
6 & 2 & marie cu \#\#rie was born in poland [MASK] [MASK] [MASK] [MASK] died in the france . & \textbf{and [PAD] [PAD] [PAD]} & \textbf{0.488} & 0.000 & 0.488 & 0.489 & 0.000 & 0.000 & 0.000 \\
6 & 3 & marie cu \#\#rie was born in poland [MASK] [MASK] [MASK] [MASK] in the france . & but but [PAD] [PAD] & 0.058 & 0.000 & 0.058 & 0.058 & 0.000 & 0.000 & 0.004 \\
6 & 4 & marie cu \#\#rie was born in poland [MASK] [MASK] [MASK] [MASK] the france . & but but up [PAD] & 0.018 & 0.000 & 0.018 & 0.018 & 0.000 & 0.000 & 0.000 \\
7 & 0 & marie cu \#\#rie was born in poland . [MASK] [MASK] [MASK] [MASK] she died in the france . & [PAD] [PAD] [PAD] [PAD] & 0.000 & 0.000 & 0.000 & 0.945 & 0.945 & 0.906 & 0.906 \\
7 & 1 & marie cu \#\#rie was born in poland . [MASK] [MASK] [MASK] [MASK] died in the france . & her [PAD] [PAD] [PAD] & 0.005 & 0.000 & 0.005 & 0.125 & 0.120 & 0.000 & 0.000 \\
7 & 2 & marie cu \#\#rie was born in poland . [MASK] [MASK] [MASK] [MASK] in the france . & her family was [PAD] & 0.010 & -0.000 & 0.010 & 0.010 & 0.000 & 0.000 & 0.000 \\
7 & 3 & marie cu \#\#rie was born in poland . [MASK] [MASK] [MASK] [MASK] the france . & she family to [PAD] & 0.008 & -0.001 & 0.007 & 0.008 & 0.000 & 0.001 & 0.000 \\
7 & 4 & marie cu \#\#rie was born in poland . [MASK] [MASK] [MASK] [MASK] france . & she was raised in & 0.011 & -0.000 & 0.011 & 0.011 & 0.000 & 0.000 & 0.000 \\
8 & 0 & marie cu \#\#rie was born in poland . she [MASK] [MASK] [MASK] [MASK] died in the france . & [PAD] [PAD] [PAD] [PAD] & 0.000 & 0.000 & 0.000 & 0.890 & 0.890 & 0.733 & 0.733 \\
8 & 1 & marie cu \#\#rie was born in poland . she [MASK] [MASK] [MASK] [MASK] in the france . & was up [PAD] [PAD] & 0.013 & -0.019 & -0.006 & 0.017 & 0.004 & 0.024 & 0.005 \\
8 & 2 & marie cu \#\#rie was born in poland . she [MASK] [MASK] [MASK] [MASK] the france . & was to [PAD] [PAD] & 0.006 & -0.003 & 0.003 & 0.006 & 0.000 & 0.003 & 0.000 \\
8 & 3 & marie cu \#\#rie was born in poland . she [MASK] [MASK] [MASK] [MASK] france . & was lives in [PAD] & 0.006 & -0.001 & 0.005 & 0.006 & 0.000 & 0.001 & 0.000 \\
8 & 4 & marie cu \#\#rie was born in poland . she [MASK] [MASK] [MASK] [MASK] . & is of in [PAD] & 0.004 & -0.002 & 0.002 & 0.004 & 0.000 & 0.002 & 0.000 \\
9 & 0 & marie cu \#\#rie was born in poland . she died [MASK] [MASK] [MASK] [MASK] in the france . & [PAD] [PAD] [PAD] [PAD] & 0.000 & 0.000 & 0.000 & 0.556 & 0.556 & 0.722 & 0.722 \\
9 & 1 & marie cu \#\#rie was born in poland . she died [MASK] [MASK] [MASK] [MASK] the france . & in [PAD] [PAD] [PAD] & 0.000 & 0.000 & 0.000 & 0.480 & 0.480 & 0.621 & 0.621 \\
9 & 2 & marie cu \#\#rie was born in poland . she died [MASK] [MASK] [MASK] [MASK] france . & in [PAD] [PAD] [PAD] & 0.504 & -0.394 & 0.110 & 0.511 & 0.007 & 0.400 & 0.007 \\
9 & 3 & marie cu \#\#rie was born in poland . she died [MASK] [MASK] [MASK] [MASK] . & in paris [PAD] [PAD] & 0.069 & -0.019 & 0.050 & 0.069 & 0.000 & 0.020 & 0.001 \\
10 & 0 & marie cu \#\#rie was born in poland . she died in [MASK] [MASK] [MASK] [MASK] the france . & [PAD] [PAD] [PAD] [PAD] & 0.000 & 0.000 & 0.000 & 0.351 & 0.351 & 0.721 & 0.721 \\
10 & 1 & marie cu \#\#rie was born in poland . she died in [MASK] [MASK] [MASK] [MASK] france . & [PAD] [PAD] [PAD] [PAD] & 0.456 & -0.442 & 0.014 & 0.462 & 0.007 & 0.449 & 0.007 \\
10 & 2 & marie cu \#\#rie was born in poland . she died in [MASK] [MASK] [MASK] [MASK] . & paris [PAD] [PAD] [PAD] & 0.091 & -0.017 & 0.074 & 0.091 & 0.000 & 0.018 & 0.001 \\
11 & 0 & marie cu \#\#rie was born in poland . she died in the [MASK] [MASK] [MASK] [MASK] france . & netherlands of [PAD] [PAD] & 0.022 & -0.113 & -0.091 & 0.022 & 0.000 & 0.113 & 0.000 \\
11 & 1 & marie cu \#\#rie was born in poland . she died in the [MASK] [MASK] [MASK] [MASK] . & netherlands states [PAD] [PAD] & 0.105 & -0.118 & -0.013 & 0.106 & 0.001 & 0.120 & 0.002 \\
12 & 0 & marie cu \#\#rie was born in poland . she died in the france [MASK] [MASK] [MASK] [MASK] . & [PAD] [PAD] [PAD] [PAD] & 0.000 & 0.000 & 0.000 & 0.223 & 0.223 & 0.437 & 0.437 \\
13 & 0 & marie cu \#\#rie was born in poland . she died in the france . [MASK] [MASK] [MASK] [MASK] & [PAD] [PAD] [PAD] [PAD] & 0.000 & 0.000 & 0.000 & 1.000 & 1.000 & 1.000 & 1.000 \\
\bottomrule
\end{tabular}
}
\caption{\label{tab:fig1} The masked inputs and the scores computed by \masker{} for the example shown in Figure 1 of the main paper to find the best span to edit to fuse the two input sentences. The last four columns show the following likelihoods: $\mathcal{L}_1 = \pl{\widehat{W}_{i:j}^\textmd{target} \mid W_{\backslash i:j}; \Theta_\textmd{target}}$, $\mathcal{L}_2 = \pl{W_{i:j} \mid W_{\backslash i:j}; \Theta_\textmd{target}}$, $\mathcal{L}_3 = \pl{\widehat{W}_{i:j}^\textmd{target} \mid W_{\backslash i:j}; \Theta_\textmd{source}}$, $\mathcal{L}_4 = \pl{W_{i:j} \mid W_{\backslash i:j}; \Theta_\textmd{source}}$.}
\end{sidewaystable*}